# Self and Mixed Supervision to Improve Training Labels for Multi-Class Medical Image Segmentation

Jianfei Liu[a], Christopher Parnell[b], Ronald M. Summers[a]

[a]Imaging Biomarkers and Computer-Aided Diagnosis Laboratory, Radiology and Imaging Sciences, National Institutes of Health Clinical Center, Bethesda, MD USA 20892
[b]Diagnostic Radiology, Walter Reed National Military Medical Center, Bethesda, MD USA 20889

## ABSTRACT

Accurate training labels are a key component for multi-class medical image segmentation. Their annotation is costly and time-consuming because it requires domain expertise. In our previous work, a dual-branch network was developed to segment single-class edematous adipose tissue. Its inputs include a few strong labels from manual annotation and many inaccurate weak labels from existing segmentation methods. The dual-branch network consists of a shared encoder and two decoders to process weak and strong labels. Self-supervision iteratively updates weak labels during the training process. This work aims to follow this strategy and automatically improve training labels for multi-class image segmentation. Instead of using weak and strong labels to only train the network once in the previous work, transfer learning is used to train the network and improve weak labels sequentially. The dual-branch network is first trained by weak labels alone to initialize model parameters. After the network is stabilized, the shared encoder is frozen, and strong and weak decoders are fine-tuned by strong and weak labels together. The accuracy of weak labels is iteratively improved in the fine-tuning process. The proposed method was applied to a three-class segmentation of muscle, subcutaneous and visceral adipose tissue on abdominal CT scans. Validation results on 11 patients showed that the accuracy of training labels was statistically significantly improved, with the Dice similarity coefficient of muscle, subcutaneous and visceral adipose tissue increased from 74.2% to 91.5%, 91.2% to 95.6%, and 77.6% to 88.5%, respectively ($p<0.05$). In comparison with our earlier method, the label accuracy was also significantly improved ($p<0.05$). These experimental results suggested that the combination of the dual-branch network and transfer learning is an efficient means to improve training labels for multi-class segmentation.

## INTRODUCTION

Medical image segmentation is a fundamental task in medical image analysis by delineating a medical image into multiple meaningful regions[1]. Quantitative measurements of these segmented regions, such as organs and tumors, can facilitate subsequent diagnosis and treatments[2]. Recent development of nnU-Net could achieve high segmentation accuracy in a self-configuration manner given enough training labels[3]. The development of nnU-Net inspired a successful segmentation tool of TotalSegmentator[4]. It could accurately segment over 100 organs on CT scans. Supervised medical image segmentation has shown a great success in medical image analysis. However, the prerequisite of accurate training labels is a major bottleneck. It is a time-consuming process to create medical training labels because they require domain expertise.

One solution was to develop unsupervised segmentation models without using labeled data[5-6]. A diffusion condensation process was used to cluster similar latent features from an autoencoder, which yielded an unsupervised image segmentation by grouping similar features into segmented image regions[5]. A data cleansing strategy was developed to identify highly confident, clean labeled data. They were used to learn a domain-invariant segmentation model and directly applied to target domains[6]. Unsupervised image segmentation could achieve reasonable performance on structures with high contrast to the background, such as the optic disk in fundus images and hyperintense tumors in MRI scans[5]. However, medical image segmentation often involves structures with low-contrast boundaries, which could fail unsupervised segmentation methods. To address this issue, self-supervised learning was studied to exaggerate the feature similarity on unlabeled data through contrastive learning on pretext tasks[7]. They were auxiliary tasks that help a model to learn useful features or model weights for the downstream main tasks. For instance, a medical volume was decomposed into a set of sub-volumes. They were randomly transformed to simulate a Rubik's Cube game. A classification model was trained to identify the actually ordered volume from the randomly ordered ones[8]. The pretrained classification model was then transferred to the downstream segmentation tasks, such as OCT fluid segmentation[9] and white matter tract segmentation[10]. Self-supervised training was also used to recurrently generate pseudo-labels for the unlabeled data to augment segmentation training[11].

Semi-supervised medical image segmentation was another solution to reduce the labeling efforts. Here, only a small set of training data were labeled, while most of them were unlabeled[12]. The usage of unlabeled data to improve the segmentation accuracy was the key for semi-supervised methods. For instance, the prediction from unlabeled data were leveraged to improve segmentation accuracy through entropy minimization[13]. Their predictions were also often enhanced by using auxiliary models, such as the mean teacher[14]. Here, the predictions of unlabeled data were required to be consistent between student and teacher models. The weights of the teacher model were iteratively updated through the ones of the student model. The similar idea was applied to the cross-consistency training except that both model weights were updated in the training process[15]. Multi-scale cross-consistency training was further proposed to compare the multi-scale predictions of the unlabeled data[16]. The cross-pseudo supervision was also developed to use the pseudo labels from the predictions of the unlabeled data as the training data[7]. The recent development of semi-supervised segmentation methods was focused on a dual-branch model that contains a shared encoder and two duplicated decoders to process labeled and unlabeled data, respectively[18,19]. Our previous work also utilized the similar model to segment edematous adipose tissue with the limited number of labeled data[20].

Both unsupervised and semi-supervised segmentation methods focused on the development of the segmentation models to use a limited number of labeled data to achieve high accuracy. Active learning instead was studied to choose a small set of representative labeled data for existing segmentation models to achieve high accuracy[21]. These representative data were selected with the highest segmentation uncertainty through a query-by-committee strategy[22]. They could be also chosen from a semi-supervised segmentation framework, where unlabeled data with less prediction confidence were put into the annotation pool[23]. This work also takes the semi-supervised framework to automatically annotate the unlabeled data. They are initialized with inaccurate labels from the existing segmentation methods[24-25]. A dual-branch network is first initialized by training inaccurate labels alone in a self-supervised learning way. After that, the encoder was frozen, and the decoder was adjusted by using accurate training labels. Inaccurate labels were iteratively replaced and updated by the pseudo labels from the decoder. Experimental results on multi-class adipose tissue and muscle segmentation showed that the updated labels from proposed method was an accurate enough for downstream segmentation models, such as nnU-Net[3].

## METHODOLOGY

Fig. 1 shows the pipeline of a dual-branch network to improve multi-class training labels through self and mixed supervision. Similar to our previous work[20], the network has a shared encoder backboned with the ResNet[26], and two identical decoders to process strong and weak labels, respectively. Here, strong labels are a small set of training labels through manual annotations. Weak labels are a large amount of automatically generated labels by combining adipose tissue segmentation from a level-set method[25] and muscle segmentation from a U-Net method[26]. Subcutaneous adipose tissue is often mislabeled as the visceral adipose tissue and pelvis muscle is also missed in the weak labels. The mixed supervision of strong and weak labels is used to train the dual-branch network.

However, the highly imbalanced multi-class training data could cause the over-fitting of the dual-branch network. This work adapts a self-supervised learning strategy to address this issue[7,277]. Weak labels are used to train the dual-branch network on the pretext task of segmenting adipose tissues and muscle because segmentation labels are available. Only the supervised data term in our earlier work[20] is used, while the cross-consistent training term and the prediction confidence term are excluded because they are designed to constrain the consistency between two decoders. The self-supervised training mainly focuses on the feature representation of the encoder. The segmentation accuracy could be improved in the downstream task. The generalized Dice overlap is used to formulate the supervised data term[28].

After self-supervised learning is complete (for 100 epochs), transfer learning is used to froze the encoder and finetune two decoders by using both strong and weak labels. Supervised data term, cross-consistent training term, and the prediction confidence term are all used to guide the finetuning process[20].

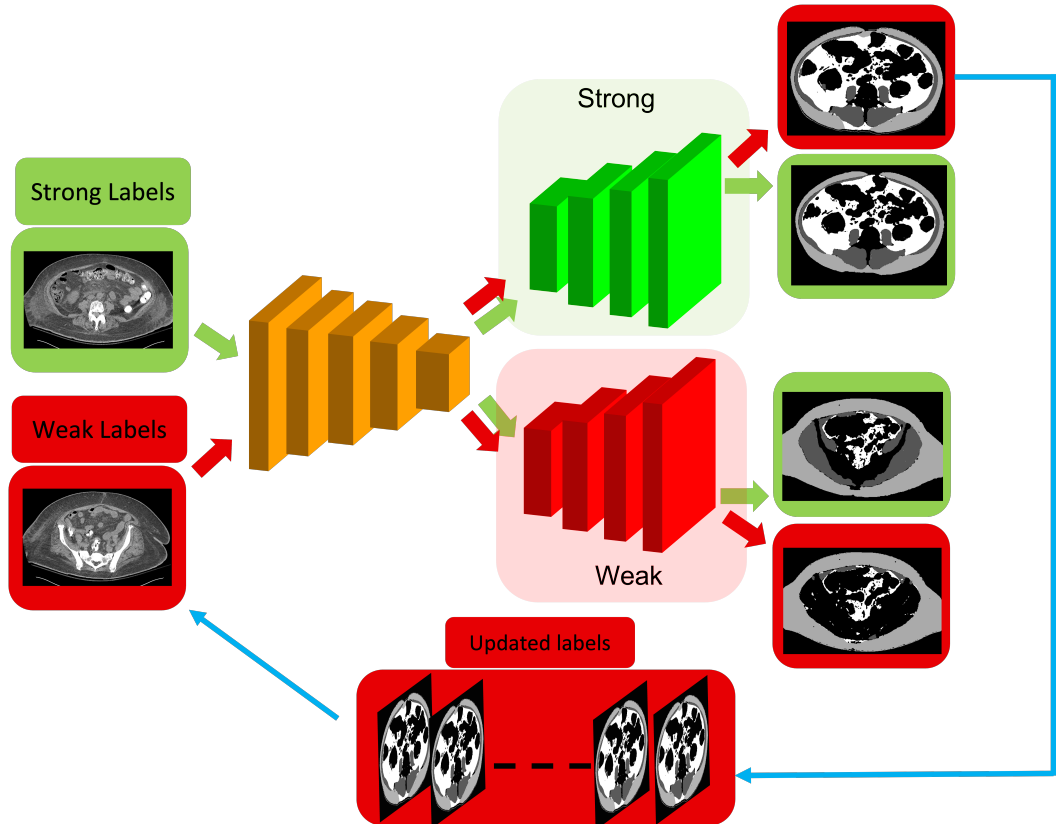

**Figure 1**: Process of multi-class training label improvement through self and mixed supervision using a dual-branch network. It consists of a shared encoder (brown) and two identical decoders to process strong labels from manual annotation (green) and inaccurate weak labels from automated segmentation methods (red). This is called mixed supervision. Self-supervision (cyan path) iteratively improves multi-class weak labels by replacing them with the segmentation masks from the strong decoder.

During the fine-tuning process, weak labels are iteratively replaced with their predicted segmentation masks from the strong decoder (Fig. 1). Fig. 2 illustrates an example of training label improvement. The missed muscle regions in the original weak labels at epoch 0 are gradually recovered at epoch 100.

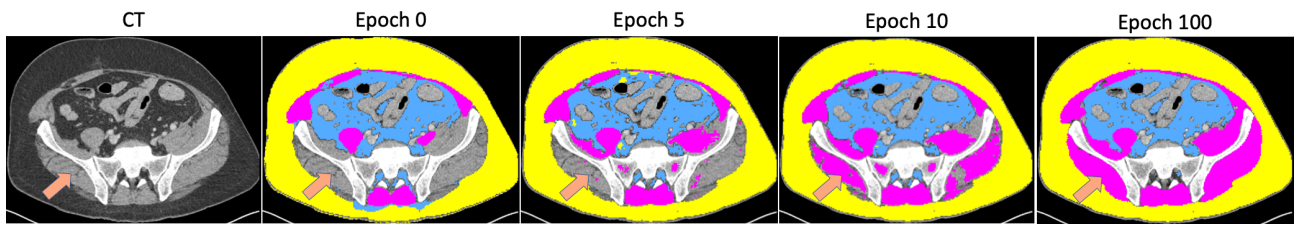

**Figure 2**: Example of multi-class training label improvement through a self-supervision process. Here, the weak label of muscle (pink) at training epoch=0 misses large regions (brown arrow). Some spotted labels appear at epoch 5, the labeled regions expand at epoch 10, and all muscle regions are labeled at epoch 100. Subcutaneous (yellow) and visceral (blue) adipose tissue labels are also enhanced by the self-supervision process.

The dataset consists of 101 abdominal CT scans from 101 patients (52 females and 49 males, average age 66.6±5.1 years) at the National Institutes of Health Clinical Center. Of these, 31 scans were randomly selected from the whole dataset. five slices from chest to pelvis were used for manual annotation in each selected scan. All 155 CT slices were manually annotated by a radiology resident, supervised by an experienced radiologist. ITK-SNAP software was used to manually annotate adipose tissue and muscle[29]. 100 of the 155 slices were used as strong labels. The remaining 55 slices were for validating the accuracy improvement of training labels. 11,326 unannotated CT slices from 101 CT scans were used as the weak labels. To evaluate accuracy, three segmentation metrics were used: intersection over union (IoU), Dice Similarity Coefficient (DSC), and relative volume difference (RVD). The training strategy in an earlier work was used as the baseline to evaluate the efficacy of transfer learning[20].

## EXPERIMENTAL RESULTS

The proposed method performed better than the baseline method (Table 1). DSC was significantly improved by 17% for muscle ($p < 0.05$), 4% for subcutaneous adipose tissue ($p < 0.05$), and 11% for visceral adipose tissue ($p < 0.05$). In comparison with the baseline, transfer learning also improved label accuracy; DSC significantly improved by 3% for muscle ($p < 0.05$), 1% for subcutaneous tissue ($p < 0.05$), and 5% for visceral tissue ($p < 0.05$).

**Table 1**: Comparison of multi-class segmentation accuracy on 11 abdominal CT scans.

| \multicolumn{4}{c}{A: Muscle} |
|---|---|---|---|
| Methods | IoU (%) | DSC (%) | RVD (%) |
| Initial Weak labels | 62.5±23.0 | 74.2±20.1 | 28.0±27.3 |
| Dual branches without transfer learning | 78.8±7.2 | 88.0±4.6 | 7.1±6.8 |
| Dual Branches with transfer learning | **84.5±4.6** | **91.5±2.7** | **4.7±5.0** |
| \multicolumn{4}{c}{B: Subcutaneous adipose tissue} |
| Methods | IoU (%) | DSC (%) | RVD (%) |
| Initial weak labels | 84.4±9.8 | 91.2±6.4 | 7.6±9.1 |
| Dual branches without transfer learning | 90.1±5.9 | 94.7±3.4 | 3.9±4.5 |
| Dual Branches with transfer learning | **91.6±4.0** | **95.6±2.2** | **2.6±2.5** |
| \multicolumn{4}{c}{C: Visceral adipose tissue} |
| Methods | IoU (%) | DSC (%) | RVD (%) |
| Initial weak Labels | 67.2±22.8 | 77.6±21.3 | 14.1±92.5 |
| Dual branches without transfer learning | 73.2±17.3 | 83.2±13.7 | 17.8±33.1 |
| Dual Branches with transfer learning | **80.2±11.7** | **88.5±7.9** | **8.5±11.4** |

Fig. 3 showed the improvement of training labels on two patients. Parts of subcutaneous adipose tissue regions were mislabeled as visceral adipose tissue (arrow, C1). These mislabeled regions were reduced by the baseline dual-branch network without transfer learning (D1). These regions were properly segmented by the dual-branch network with transfer learning (E1). Similar results were observed in the second patient with muscle regions missed in the weak labels (arrow, 3C2). They were partly recovered by the baseline method (3D2), and completely recovered by transfer learning (3E2) in comparison with the ground truth labels (3B2).

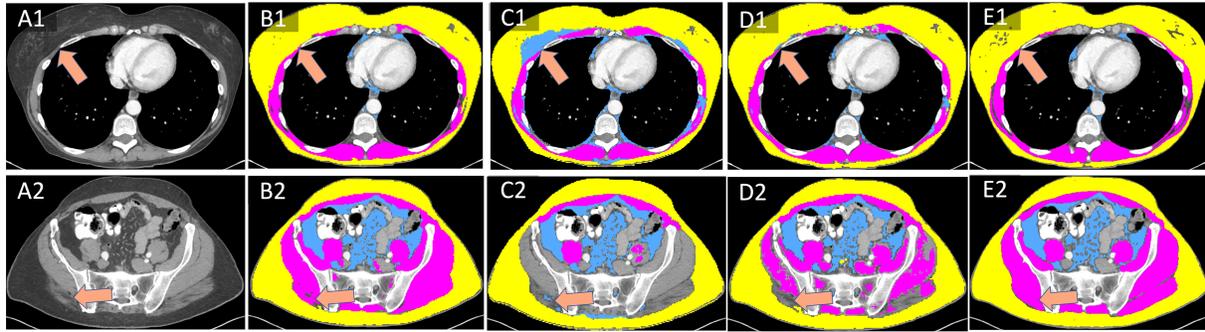

**Figure 3:** Multi-class training label improvement on (top row) abdominal and (bottom row) pelvic CT images of two patients. In comparison with ground truth labels (B1-B2), either subcutaneous adipose tissue were mislabeled as visceral adipose tissue (arrow, C1) or pelvis muscle were totally missed (arrow, C2) in the weak labels. Dual branches without transfer learning could partially improve them (D1-D2). The dual branches with transfer learning achieved the better label improvement (E1-E2).

## SUMMARY

In this paper, we presented a dual-branch network to automatically and accurately update multi-class training labels. The method only required a small number of manual annotations, while the majority of the training labels were automatically created. Self-supervised learning and transfer learning were used to train the dual-branch network and improve multi-class training labels. Experimental results on multi-class segmentation of muscle, and subcutaneous and visceral adipose tissue demonstrated the efficacy of the proposed method.

## ACKNOWLEDGEMENTS

This research was supported by the Intramural Research Program of the National Institutes of Health Clinical Center.